\documentclass[10pt,twocolumn,letterpaper]{article}

\usepackage{iccv}
\usepackage{times}
\usepackage{epsfig}
\usepackage{graphicx}
\usepackage{amsmath}
\usepackage{amssymb}
\usepackage{makecell}

\usepackage{multirow}
\usepackage{amsthm,amsmath,amssymb,lipsum}
\usepackage{mathrsfs}
\usepackage{enumerate}
\usepackage{colortbl}
\usepackage{enumitem}
\usepackage{wrapfig}
\usepackage{bbding}
\usepackage{pifont}
\usepackage{wasysym}
\usepackage{textcomp}

\usepackage{color}

\newcommand\cb[1]{\color{blue} #1}
\newcommand\cred[1]{\color{red} #1}
\newcommand{\cmark}{\ding{51}}%
\newcommand{\xmark}{\ding{55}}%
\definecolor{Gray}{gray}{0.95}
\definecolor{Gray7}{gray}{0.75}

\usepackage{microtype}
\usepackage{graphicx}
\usepackage{booktabs} 
\usepackage{adjustbox}
\usepackage{subcaption}

\usepackage[breaklinks=true,bookmarks=false]{hyperref}

\iccvfinalcopy 

\def\iccvPaperID{5236} 

\begin{document}

\title{Take-A-Photo: 3D-to-2D Generative Pre-training of Point Cloud Models}

\author{
    Ziyi Wang\thanks{Equal contribution. ~\textsuperscript{\dag}Corresponding author.} ~~~~
  	Xumin Yu$^*$ ~~
	Yongming Rao ~~
	Jie Zhou ~~~
	Jiwen Lu$^{\dagger}$      \\
    Department of Automation, Tsinghua University, China\\
    {\tt\small \{wziyi22, yuxm20\}@mails.tsinghua.edu.cn;} \\{\tt\small raoyongming95@gmail.com; \{jzhou, lujiwen\}@tsinghua.edu.cn} \\
}

\maketitle
\ificcvfinal\thispagestyle{empty}\fi

\begin{abstract}
   With the overwhelming trend of mask image modeling led by MAE, generative pre-training has shown a remarkable potential to boost the performance of fundamental models in 2D vision. However, in 3D vision, the over-reliance on Transformer-based backbones and the unordered nature of point clouds have restricted the further development of generative pre-training. In this paper, we propose a novel 3D-to-2D generative pre-training method that is adaptable to any point cloud model. We propose to generate view images from different instructed poses via the cross-attention mechanism as the pre-training scheme. Generating view images has more precise supervision than its point cloud counterpart, thus assisting 3D backbones to have a finer comprehension of the geometrical structure and stereoscopic relations of the point cloud. Experimental results have proved the superiority of our proposed 3D-to-2D generative pre-training over previous pre-training methods. Our method is also effective in boosting the performance of architecture-oriented approaches, achieving state-of-the-art performance when fine-tuning on ScanObjectNN classification and ShapeNetPart segmentation tasks. Code is available at \url{https://github.com/wangzy22/TAP}.
\end{abstract}

\section{Introduction}

Nowadays, pre-training fundamental models with self-supervised mechanisms has witnessed a thriving development in the computer vision community, given its low requirement in data annotation and its high transferability to downstream applications. Self-supervised pre-training aims to fully exploit the statistical and structural knowledge from abundant annotation-free data and empowers the fundamental models with robust representation ability. In 2D vision, self-supervised pre-training has shown strong potential and achieved remarkable progress on diverse backbones in various downstream tasks. Successful pre-training strategies in 2D domain such as contrastive learning~\cite{moco, SimCLR} and mask modeling~\cite{mae, beit} have also been adopted to 3D point cloud analysis~\cite{yu2022point, pang2022masked, liu2022masked} in recent research.

\begin{figure}[t!]
    \begin{center}
    \includegraphics[width=\linewidth]{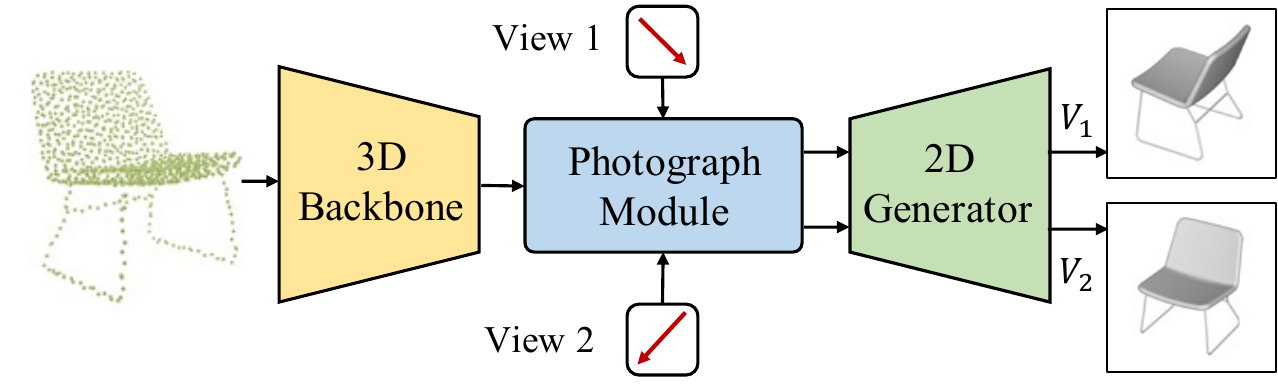}
    \caption{\textbf{Principle illustration of 3D-to-2D generative pre-training.} The photograph module explicitly encodes pose condition into 3D features from the backbone, and the 2D generator decodes pose-conditioned features into different view images.}\label{fig:concept}
    \end{center}
    \vspace{-6pt}
\end{figure}

However, the pre-training paradigm hasn't become the prevailing trend in 3D learning and architectural design is still the mainstream to reach a new state-of-the-art performance, which is considerably different from the dominant status of pre-training in 2D domain. In object-level analysis, generative pre-training inspired by MAE~\cite{mae} has been studied thoroughly, but their performances still lag behind architecture-based methods like PointNeXt~\cite{pointnext}. Two factors mainly contribute to the inferior status of generative pre-training in 3D learning. Since point clouds are unordered sets of point coordinates, there is no direct and precise supervision like one-to-one MSE loss between generated point clouds and their corresponding ground truth. Chamfer Distance supervision for point clouds only calculates a rough set-to-set matching and its imprecision has been widely discussed in~\cite{liu2020morphing, wu2021balanced, huang2023learning}. Additionally, existing advanced generative pre-training methods in object analysis are limited to the Transformer-based backbone, and fail to be extended to other sophisticated point cloud models.

To alleviate the aforementioned problems, we propose a 3D-to-2D generative pre-training method for point cloud analysis that has higher preciseness in supervision and broader adaptation to different backbones. Instead of reconstructing point clouds as previous literature~\cite{pang2022masked, pointm2ae}, we propose to generate view images of the input point cloud given the instructed camera poses. This is similar to taking photos of a realistic object from different perspectives to fully present its structure or internal relations. Therefore, we name our model \textit{Take-A-Photo}, in short \textit{TAP}. More specifically, we propose a pose-dependent photograph module that utilizes the cross-attention mechanism to explicitly encode pose conditions with 3D features from the backbone. Then a 2D generator decodes pose-conditioned features into view images that are supervised by rendered ground truth images. The principle illustration of TAP is shown in Figure~\ref{fig:concept}. In the pose-dependent photograph module, we do not provide detailed projection relations from 3D points to 2D pixels, thus the cross-attention layers are encouraged to learn by themselves what those view images look like conditioned on given poses. Since the projection layout, part occlusion relation, faces colors that represent light reflections are largely distinct among view images, the proposed 3D-to-2D generative pre-training is a challenging pretext task that obliges 3D backbone to gain higher representation ability of geometrical knowledge and stereoscopic relations.

We conduct extensive experiments on various backbones and downstream tasks to verify the effectiveness and superiority of our proposed 3D-to-2D generative pre-training method. When pre-trained on synthetic ShapeNet~\cite{shapenet} and transferred to real-world ScanObjectNN~\cite{uy2019revisiting} classification, TAP brings consistent improvement to different backbone models and successfully outperforms previous point cloud generative pre-training methods based on Transformers backbone. With PointMLP~\cite{pointmlp} as the backbone, TAP achieves state-of-the-art performance on ScanObjectNN classification and ShapeNetPart~\cite{shapenetpart} part segmentation among methods that do not include any pre-trained image or text model. We also conduct thorough ablation studies to discuss the architectural design of the TAP model and verify the individual contribution of each component. 

In conclusion, the contributions of our paper can be summarized as follows:
(1) We propose TAP, the first 3D-to-2D generative pre-training method that is adaptable to any point cloud model. TAP pre-training helps to exploit the potential of point cloud models on geometric structure comprehension and stereoscopic relation understanding. (2) We propose a Photograph Module where we derive mathematical formulations to encode pose conditions as query tokens in cross-attention layers. (3) TAP surpasses previous generative pre-training methods on the Transformers backbone and achieves state-of-the-art performance on ScanObjectNN classification and ShapeNetPart segmentation.

\section{Related Work}

\begin{figure*}[t]
    \begin{center}
    \includegraphics[width=\linewidth]{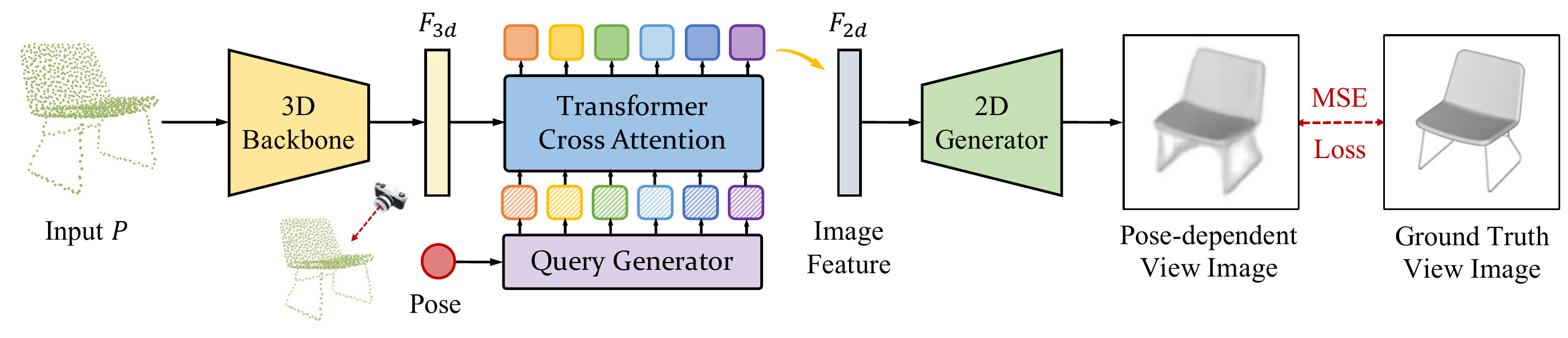}
    \caption{\textbf{The pipeline of TAP pre-training method.} We design a query generator to encode pose conditions and implement cross-attention layers to transform 3D point cloud features $F_\textrm{3d}$ to 2D view image features $F_\textrm{2d}$ according to pose instruction. The predicted pose-dependent view image from 2D generator is supervised by ground truth view image via MSE loss.}\label{fig:pipeline}
    \end{center}
\end{figure*}

\noindent\textbf{Point Cloud Analysis.} 
Point cloud analysis is a fundamental and important task in the realm of 3D vision. Current literature has developed two principal methodologies to extract structural representations from 3D point clouds, namely point-based and voxel-based methods. Point-based methods~\cite{pointnet, pointnet2, wang2019dynamic, thomas2019KPConv, pointnext, pointmlp} process unordered points directly, introducing various techniques for local information aggregation. Existing point-based methods can be categorized into three types: SetAbstraction-based~\cite{pointnet, pointnet2, pointnext}, DynamicGraph-based~\cite{wang2019dynamic}, and Attention-based~\cite{yu2021pointr, yu2022point, pang2022masked, liu2022masked, pointm2ae}, all of which focus on modeling the relationships between points. 
Owing to their exceptional ability to effectively preserve fine-grained geometric information, point-based methods are frequently employed for object-level tasks. 
On the other hand, voxel-based methods~\cite{maturana2015voxnet, klokov2017escape, riegler2017octnet} partition the 3D space into ordered voxels and employ 3D convolutions for feature extraction. Voxel-based methods are primarily based on SparseConvolution~\cite{choy2019minkowski, graham2018sparseconv}, which enables efficient convolution operations in 3D space through sparse convolutions. 
In exchange for faster processing speeds, voxel-based methods sacrifice their capacity to capture detailed local structures, making them more suitable for large-scale scene-level tasks rather than object-level tasks. 

\vspace{6pt}
\noindent\textbf{Point Cloud Pre-training.}
Pre-training has always been a focal point of research in the field of deep learning. Generally speaking, we usually distinguish pre-training methods based on the amount of annotation required, namely full-supervised pre-training~\cite{dosovitskiy2020vit, zhai2022scaling, carreira2017quo}, weakly-supervised pre-training~\cite{tarvainen2017mean, berthelot2019mixmatch, xie2020self}, and unsupervised pre-training~\cite{moco, SimCLR, chen2020improved}. Among these methods, unsupervised pre-training has become the most popular approach, mainly due to its excellent transferability across tasks and its notable advantage of not relying on labeled data. Numerous researchers have proposed a variety of pretext tasks for unsupervised pre-training. 
Based on the type of pretext task employed, there are two prevailing pretext tasks for unsupervised pre-training. The first is contrastive learning, as exemplified by MoCo~\cite{moco} and SimCLR~\cite{SimCLR}. The other method entails utilizing generative tasks to restore the data from partially or disrupted inputs, such as MAE~\cite{mae} and BEiT~\cite{beit}.

Inspired by pre-training strategies in the image domain, there are more and more unsupervised pre-training methods being proposed for point cloud pre-training. PointContrast~\cite{pointcontrast} embraces the principle of contrastive learning, whereas PointBERT~\cite{yu2022point} and PointMAE~\cite{pang2022masked} integrate reconstruction pretext tasks. However, existing generative-based pre-training methods for point clouds only consider a single modality. In this paper, we propose a cross-modal generative-based pre-training strategy to achieve more effective pre-training.

\vspace{6pt}
\noindent\textbf{Cross-Modal Learning.}
Recently, cross-modal learning has been a popular research topic, aiming at learning from multiple modalities such as images, audio and point clouds. It has the potential to enhance the performance of various tasks, including visual recognition, speech recognition, and point cloud analysis. A variety of methods have been proposed for cross-modal learning, including multi-task learning~\cite{collobert2008unified}, conditional generation~\cite{rombach2022high}, pre-training~\cite{qi2023recon, dong2022act} and tuning~\cite{wang2022p2p, xu2021image2point}.

In the realm of point cloud analysis, much previous work has explored this learning paradigm. Some literature leverages 2D data for 3D point cloud analysis, such as MVTN~\cite{MVTN}, MVCNN~\cite{su2015multi} and CrossPoint~\cite{crosspoint}, proving that the multi-view images and the correspondence between images and points can be helpful for the 3D object understanding. Another line of the research, such as Image2Point~\cite{xu2021image2point} and P2P~\cite{wang2022p2p} take effort to adapt the models from 2D vision into 3D point cloud analysis, fully exploiting the relationship of 2D and 3D understanding. In this paper, we continue this learning paradigm in 3D vision domain, and for the first time propose the cross-modal generative pre-training scheme for point cloud models.
\section{3D-to-2D Generative Pre-training}
\subsection{Preliminary: Generative Pre-training}

Generative pre-training is a fundamental branch of pre-training methods that aims at reconstructing integral and complete data given partial or disrupted input. Mathematically, suppose $x$ is a sample from raw data with no annotation. The pre-processing step $T(\cdot)$ either erases part of $x$ randomly or splits $x$ into pieces and intermingles them to get $\tilde{x}=T(x)$. The generative pre-training model $M$ is designed to restore from those broken input $\hat{x}=M(T(x))$ and the training loss function is designed to measure the reconstruction distance $\mathcal{L}=D(\hat{x}, x)$.
In point cloud object analysis, earlier generative pre-training methods propose various pretext tasks as $T$, including deformation~\cite{achituve2021self}, jigsaw puzzles~\cite{Jigsaw3D} and depth projection~\cite{occo} to produce disarrayed or partial point clouds. Recently, inspired by MAE~\cite{mae} in the image domain, generative pre-training in 3D domain mainly focuses on implementing random masking as $T$ and utilizing Transformers model as $M$ for reconstruction~\cite{yu2022point, pang2022masked, liu2022masked, pointm2ae}. The reconstruction distance $D$ is usually measured by the classical $l_2$ Chamfer Distance:
\begin{equation}
    D(\hat{x},x)=\frac{1}{\lvert \hat{x}\rvert}\sum_{a\in \hat{x}}\min_{b\in x}\lVert a-b \rVert_2^2 + \frac{1}{\lvert x\rvert}\sum_{b\in x}\min_{a\in \hat{x}}\lVert a-b \rVert_2^2
\label{eq:chamfer}
\end{equation}
Besides Chamfer Distance between point clouds, some methods also exploit feature distance between latents~\cite{yu2022point} or occupancy value distance~\cite{liu2022masked} as the loss function. 

The exact reason why generative pre-training would help enhance the representation ability of backbone models still remains an open question. However, abundant experimental results have conveyed that predicting missing parts according to known parts demands high reasoning ability and global comprehension capacity of the model. What's more, generative pre-training is more efficient and suitable for point cloud object analysis than contrastive pre-training, given that contrastive pre-training typically requires a large amount of training data to avoid trivial overfitting solutions but there has always been a data-starvation problem in point cloud object research field. 

\subsection{Overall Pipeline}

Different from the aforementioned generative pre-training methods that focus on uni-modal point cloud reconstruction, we propose a novel cross-modal pre-training approach of generating view images from instructed camera poses. 

The overall architecture of our proposed TAP pre-training model is depicted in Figure~\ref{fig:pipeline}. Our model takes as an input point cloud $P\in \mathbb{R}^{N\times 3}$, where $N$ is the number of points in the input point cloud. The basic building block of TAP mainly consists of: 1) a \textit{3D Backbone} that extracts 3D geometric features $F_\textrm{3d}\in \mathbb{R}^{n\times C_\textrm{3d}}$, where $n$ is the number of downsampled center points and $C_\textrm{3d}$ is the geometric feature dimension; 2) a \textit{pose-dependent Photograph Module} that takes as inputs $F_\textrm{3d}$ and pose matrix $R\in \mathbb{R}^{3\times 3}$, and predicts view image features $F_\textrm{2d}^R\in \mathbb{R}^{h\times w\times C_\textrm{2d}}$ conditioned on $R$, where $h, w$ are height and width of predicted view image feature map; 3) an \textit{2D Generator} that decodes $F_\textrm{2d}^R$ into an RGB image $I^R_\textrm{gen}\in \mathbb{R}^{H\times W\times 3}$, where $H, W$ are height and width of the output view image. 

As we place no restriction on $F_\textrm{3d}$, the \textit{3D Backbone} can be arbitrarily chosen and adopted. Therefore, our TAP is more flexible and compatible than existing generative pre-training methods that are limited to Transformer-based architecture. Experimental results in Section~\ref{sec:exp} will later verify that TAP brings consistent improvement to all kinds of point cloud models. The technical designs of the \textit{pose-dependent Photograph Module} will be thoroughly discussed in Section~\ref{sec:photo_module}. The \textit{2D Generator} consists of four Transpose Convolution layers to progressively upsample image resolution and decode RGB colors of each pixel.

\subsection{Photograph Module}
\label{sec:photo_module}

\noindent\textbf{Architectural Design.} As illustrated in Figure~\ref{fig:pipeline}, we leverage cross-attention mechanism from Transformers~\cite{vaswani2017attention} to build our \textit{pose-dependent Photograph Module}.
\begin{equation}
    \textrm{Attention}(Q,K,V) = \textrm{softmax}\left(\frac{QK^T}{\sqrt{d_k}}\right)V
\end{equation}
where $d_k$ is the scaling factor, and $Q,K,V$ are quries, keys and values matrix. More specifically, we design a Query Generator $\Phi$ to encode camera pose conditions into query tokens: $Q=\Phi(R)\in \mathbb{R}^{hw\times C_\textrm{2d}}$. We also design a Memory Builder $\Theta$ to construct $K$ and $V$ from 3D geometric features: $K=V=\Theta(F_\textrm{3d})\in \mathbb{R}^{m\times C_\textrm{2d}}$, where $m$ is the number of memory tokens. The output sequence of the cross attention layers will be rearranged from $hw \times C_\textrm{2d}$ to $h\times w \times C_\textrm{2d}$, forming the predicted view image features $F_\textrm{2d}^R$.

During the cross-attention calculation process, we do not explicitly provide any projection clues of which 3D points would project to which 2D pixel. Instead, the Photograph Module learns by itself how to arrange unordered 3D feature points to ordered 2D pixel grids, purely based on semantic similarities between 3D geometric features and our delicately-designed queries that reveal pose information. Since one sample will only have one set of memory tokens in 3D space but its view images from different poses are quite distinct from each other, learning to predict precise view images from instructed poses in a data-driven manner is not a trivial task. Therefore, during the end-to-end optimization process, the 3D backbone is trained to have a stronger perception of the object's overall geometric structure and gain a higher representative ability of the stereoscopic relations. In this way, our proposed 3D-to-2D generative pre-training would help exploit the potential and enhance the strength of 3D backbone models.

\vspace{6pt}
\noindent\textbf{Query Generator.} The query generator $\Phi$ is designed to encode pose condition $R$ into 2D grid of shape $h\times w$. In object analysis, common practice is leveraging parallel light shading to project 3D objects onto 2D grids, and pose matrix $R$ here is used to rotate objects into desired angles before projection. Therefore, each 2D grid actually represents an optical line that starts from infinity, passes through 3D objects and ends at the 2D plane. As a consequence, we choose the direction and the origin points that the optical line goes through as the delegate of the query grid. 

Before deriving formulations of optical lines for each grid, let us first revisit the parallel light shading process for better comprehension. Given 3D coordinates $\mathbf{x}=(x,y,z)$ of a point cloud $P$ and pose matrix $R$, rotation is first performed to align the object to the ideal pose position:
\begin{equation}
    \mathbf{x'}=(x',y',z')=R\mathbf{x}
\label{eq:rotate}
\end{equation}
Then we just omit the final dimension $z'$ and evenly split the first two dimensions $(x',y')$ into 2D grids $(u,v)$:
\begin{equation}
\begin{aligned}
    u = \frac{x'-x_0}{g_h} + o_h, \quad
    v = \frac{y'-y_0}{g_w} + o_w
\label{eq:proj}
\end{aligned}
\end{equation}
where $(x_0, y_0)$ is the minimum value of $(x',y')$, $(g_h, g_w)$ is the grid size, $(o_h, o_w)$ is the offset value to place the projected object at the center of the image. $0\leq u \leq h-1, 0\leq v \leq w-1$ and $(u,v)$ is a sampled pixel coordinate from the 2D grid.

Now let us begin to derive formulations of the optical line that passes through the query grid. We only know $(u,v)$ for each grid and we want to reversely trace which 3D points $(x,y,z)$ are on the same optical line during parallel light projection. According to Eq.~\ref{eq:proj}:
\begin{equation}
\begin{aligned}
    x' &= g_h u + x_0 - o_h = \Psi_h(u) \\
    y' &= g_w v + y_0 - o_w = \Psi_w(v)
\end{aligned}
\end{equation}
If we denote $A=R^{-1}$ and $A_{ij}$ as the element at $i^{th}$ row and $j^{th}$ column, then according to Eq.~\ref{eq:rotate}:
\begin{equation}
\begin{aligned}
    x &= A_{11}\Psi_h(u) + A_{12}\Psi_w(v) + A_{13}z' = \Omega_x(u,v) + A_{13}z' \\
    y &= A_{21}\Psi_h(u) + A_{22}\Psi_w(v) + A_{23}z' = \Omega_y(u,v) + A_{23}z' \\
    z &= A_{31}\Psi_h(u) + A_{32}\Psi_w(v) + A_{33}z' = \Omega_z(u,v) + A_{33}z' 
\label{eq:line}
\end{aligned}   
\end{equation}
According to the definition of line's parametric equation, Eq.~\ref{eq:line} represents a line passing through the origin point $O:(\Omega_x(u,v), \Omega_y(u,v), \Omega_z(u,v))$ with optical line direction $\mathbf{d}=(A_{13}, A_{23}, A_{33})$, where $\Omega_x, \Omega_y, \Omega_z$ are $xyz$ coordinates of $O$ and their formulations are conditioned on $u,v$. Therefore, we concatenate the coordinate of origin point $O$, normalized direction $\mathbf{d}^\dagger = \mathbf{d} / \lVert \mathbf{d} \rVert_2$ and normalized position $(u/h,v/w)$ as positional embedding together to be the initial state of our query. A multi-layer-perceptron (MLP) module is later leveraged to map the 8-dim initial query to higher dimensional space.

\vspace{6pt}
\noindent\textbf{Memory Builder.} The memory builder takes $F_\textrm{3d}$ as input to prepare for initial state of $K, V$ in cross-attention layers. We first concatenate aligned 3D coordinate $P_\textrm{3d}$ with 3D features to enhance the geometric knowledge of $F_\textrm{3d}$:
\begin{equation}
    \hat{F}_\textrm{3d} = \mathrm{MLP}(\mathrm{cat}(F_\textrm{3d}, P_\textrm{3d}))
\end{equation}
Additionally, we initialize a learnable memory token $T_\textrm{pad}$ as the pad token and concatenate it with $\hat{F}_\textrm{3d}$ to obtain the initial state of $K, V$. The reason for concatenating a learnable pad token $T_\textrm{pad}$ is that there are white background areas on the projected image (as shown in Figure~\ref{fig:pipeline}). As $F_\textrm{3d}$ only encodes foreground objects, we further need a learnable pad token to represent background regions. Otherwise, the cross-attention layers will be confused to learn how to combine foreground tokens into background features and this will inevitably diminish the pre-training effectiveness.

\subsection{Objective Function}

We perform per-pixel supervision with Mean Squared Error (MSE) loss between generated view image $I^R_\textrm{gen}$ and ground truth image $I^R_\textrm{gt}$, aligned by camera pose $R$. For simplicity, we will omit $R$ in later formulations. As the background of the rendered ground truth images is all white and reveals little information, we further design a compound loss to balance the weight between foreground regions and background regions:
\begin{equation}
    \mathcal{L}(I_\textrm{gen}, I_\textrm{gt}) = w^\textrm{fg} \mathcal{D}^\textrm{fg} + w^\textrm{bg} \mathcal{D}^\textrm{bg}
\end{equation}
\begin{equation}
    \mathcal{D}^{k}(I^{k}_\textrm{gen}, I^{k}_\textrm{gt}) = \frac{1}{HW}\sum_{h,w}(I^{k}_{\textrm{gen}}(h,w) - I^{k}_{\textrm{gt}}(h,w))^2
\end{equation}
where $k=\textrm{fg (foreground)}, \textrm{bg (background)}$ and $w^\textrm{fg}, w^\textrm{bg}$ are loss weights for foreground and background, respectively. Such per-pixel supervision is more precise than the ambiguous set-to-set Chamfer Distance introduced in Eq.~\ref{eq:chamfer}. 

\section{Experiments}
\label{sec:exp}
In this section, we first introduce the setups of our pre-training scheme. Then we evaluate our pre-training method on various point cloud backbones by fine-tuning them on different downstream tasks, such as point cloud classification on ModelNet and ScanObjectNN datasets, and part segmentation on the ShapeNetPart dataset. Finally, we provide in-depth ablation studies for the architectural design of our proposed TAP pre-training pipeline.

\subsection{Pre-training Setups}
\noindent\textbf{Data Setups} 
To align with previous research practices~\cite{yu2022point, pang2022masked, pointcontrast}, we choose \textbf{ShapeNet}~\cite{shapenet} that contains more than 50 thousand CAD models as our pre-training datasets. We sampled 1024 points from each 3D CAD model to form the point clouds, consistent with previous work. Since ShapeNet does not provide images for each point cloud, we use the rendered image from 12 surrounding viewpoints generated by MVCNN~\cite{su2015multi}. During our pre-training, the models are exclusively pre-trained with the training split following the practice of previous work~\cite{yu2022point}.

\vspace{5pt}
\noindent\textbf{Architecture Setups}
We conduct experiments on various point cloud encoders, including PointNet++~\cite{pointnet2}, DGCNN~\cite{wang2019dynamic}, PointMLP~\cite{pointmlp} and Transformers~\cite{yu2022point} for point cloud object classification. During the pre-training stage, the photograph module takes encoded point cloud features and pose conditions as inputs to generate a 32$\times$ downsampled view image feature map of size $7\times 7$ from a specific viewpoint. Then the 2D generator progressively upsamples the image feature map to decode RGB view images of size $224\times 224$. We do not alter the architecture of the point cloud backbone since the photograph module and the 2D generator are exclusively used during the pre-training phase and are dropped during the fine-tuning stage. In our experiment, the photograph module is a six-layer cross-attention block, with attention layer channels limited to 256 to enhance efficiency. During the pre-training task on ShapeNet, we utilized four simple transpose convolutions to upsample the reconstructed 2D feature map and predict the RGB value for each pixel.

\begin{table}[!t]
\caption{
\textbf{Classification results on the ScanObjectNN dataset}. We report the overall accuracy (\%). The results with $\dagger$ are reproduced by PointNeXt~\cite{pointnext} repository.}
\label{tab:scanobjectnn}
\begin{center}
\vspace{-10pt}
\resizebox{1.0\linewidth}{!}{
\begin{tabular}{lccc}
\toprule[0.95pt]
Method & OBJ\_BG & OBJ\_ONLY & PB\_T50\_RS\\
\midrule[0.6pt]
\multicolumn{4}{c}{\textit{Hierarchical Models with TAP Pre-training}}\\
\midrule[0.6pt]
$^\dagger$DGCNN~\cite{wang2019dynamic} & - & - & 86.1 \\
$\quad$ + TAP & - & - & 86.6~\cred{(+0.5)} \\
$^\dagger$PointNet++~\cite{pointnet2} & - & - & 86.2 \\
$\quad$ + TAP & - & - & 86.8~\cred{(+0.6)} \\
$^\dagger$PointMLP~\cite{pointmlp} & - & - & 87.4 \\
$\quad$ + TAP & - & - & 88.5~\cred{(+1.1)} \\
\midrule[0.6pt]
\multicolumn{4}{c}{\textit{Standard Transformers with Generative Pre-training}}\\
\midrule[0.6pt]
w/o pre-training~\cite{vaswani2017attention} & 79.86 & 80.55 & 77.24 \\
OcCo~\cite{occo} & 84.85 & 85.54 & 78.79 \\
Point-BERT~\cite{yu2022point} & 87.43 & 88.12 & 83.07 \\
MaskPoint~\cite{liu2022masked} & 89.30 & 88.10 & 84.30 \\
Point-MAE~\cite{pang2022masked} & 90.02 & 88.29 & 85.18 \\
TAP (Ours) & \textbf{90.36} & \textbf{89.50} & \textbf{85.67} \\
\bottomrule[0.95pt]
\end{tabular}}
\end{center}
\vspace{-20pt}
\end{table}

\begin{table}[!t]
\caption{
\textbf{Results comparisons with previous methods on the ScanObjectNN and ModelNet40 datasets}. The model parameters number (\#Params) and overall accuracy (\%) are reported. $\dagger$ denotes our reproduced results of PointMLP with PointNeXt codebase. Methods with $*$ introduce extra knowledge from pre-trained image models or pre-trained vision-language models. We do not compete with them for fair comparison and only list them for reference.}
\label{tab:compare}
\begin{center}
\vspace{-10pt}
\renewcommand{\tabcolsep}{3pt}
\resizebox{1.0\linewidth}{!}{
\begin{tabular}{lccc}
\toprule[0.95pt]
Method & \#Params (M) & ScanObjectNN & ModelNet40 \\
\midrule[0.6pt]
\multicolumn{4}{c}{\textit{Supervised Learning Only}}\\
\midrule[0.6pt]
PointNet~\cite{pointnet} & 3.5 & 68.0 & 89.2 \\
PointNet++~\cite{pointnet2} & 1.5 & 77.9 & 90.7 \\
Transformer~\cite{vaswani2017attention} & 22.1 & 77.24 & 91.4 \\
DGCNN~\cite{wang2019dynamic} & 1.8 & 78.1 & 92.9 \\
PointCNN~\cite{li2018pointcnn} & 0.6 & 78.5 & 92.2 \\
DRNet~\cite{drnet} & - & 80.3 & 93.1\\
SimpleView~\cite{SimpleView} & - & 80.5$\pm$0.3 & 93.9 \\
GBNet~\cite{GBNet} & 8.8 & 81.0 & 93.8 \\
PRA-Net~\cite{PRANet} & 2.3 & 81.0 & 93.7 \\
MVTN~\cite{MVTN} & 11.2 & 82.8 & 93.8 \\
RepSurf-U~\cite{ran2022repsurf} & 1.5 & 84.3 & \textbf{94.4} \\
PointMLP~\cite{pointmlp} & 12.6 & 85.4$\pm$0.3 & 94.1 \\
PointNeXt~\cite{pointnext} & 1.4 & 87.7$\pm$0.4 & 93.7 \\
\midrule[0.6pt]
\multicolumn{4}{c}{\textit{Transformers with Pre-training}}\\
\midrule[0.6pt]
OcCo~\cite{occo} & 22.1 & 78.8 & 92.1 \\
Point-BERT~\cite{yu2022point} & 22.1 & 83.1 & 93.2 \\
MaskPoint~\cite{liu2022masked} & 22.1 & 84.3 & 93.8 \\
Point-MAE~\cite{pang2022masked} & 22.1 & 85.2 & 93.8 \\
Point-M2AE~\cite{pointm2ae} & 15.3 & 86.4 & 94.0 \\
\midrule[0.6pt]
\multicolumn{4}{c}{\textit{With Pre-trained Image Model}}\\
\midrule[0.6pt]
ACT~\cite{dong2022act}$^*$ & 22.1 & 88.2 & 93.7 \\
I2P-MAE~\cite{zhang2023I2PMAE}$^*$ & - & 90.1 & 93.7 \\
ReCon~\cite{qi2023recon}$^*$ & 44.3 & 90.6 & 94.1 \\
\midrule[0.6pt]
\multicolumn{4}{c}{\textit{Our Proposed TAP Pre-training}}\\
\midrule[0.6pt]
PointMLP (reproduce) & 12.6 & 87.4$^\dagger$ & 93.7$^\dagger$ \\
PointMLP + TAP & 12.6 & \textbf{88.5} & 94.0 \\
\bottomrule[0.95pt]
\end{tabular}
}
\end{center}
\vspace{-12pt}
\end{table}

\vspace{6pt}
\noindent\textbf{Implementation Details}
The experiments of TAP pre-training and finetuning on various downstream tasks are implemented with PyTorch~\cite{paszke2019pytorch}. We utilize AdamW~\cite{adamw} optimizer and the CosineAnnealing learning rate scheduler~\cite{loshchilov2016sgdr} to pre-train the point cloud backbone for 100 epochs. We set the initial learning rate as $5e^{-4}$ and weight decay as $5e^{-2}$. In our experiment, we train various point cloud backbones with 32 batch sizes on a single Nvidia 3090Ti GPU. The drop path rate of the cross-attention layer is set to 0.1. The foreground and background loss weights $w^\textrm{fg}, w^\textrm{bg}$ are set to 20 and 1. The detailed architecture of our simple 2D generator is: $\text{TConv}(256, 128, 5, 4)\rightarrow\text{TConv}(128, 64, 3, 2)\rightarrow\text{TConv}(64, 32, 3, 2)\rightarrow\text{TConv}(32, 3, 3, 2)$, where $\text{TConv}$ stands for Transpose Convolution and the four numbers in the tuple denotes $(C_{in}, C_{out}, \text{Kernel}, \text{Stride})$ respectively.
During the fine-tuning stage, we perform a learning rate warming up for point cloud backbones with 10 epochs, and keep other settings unchanged for a fair comparison.

\begin{table*}[!t]
\caption{ \textbf{Part segmentation results on the ShapeNetPart dataset}. We report the mean IoU across all part categories mIoU$_C$, the mean IoU across all instances mIoU$_I$, and the IoU for each category.}
\vspace{-5pt}
\label{tab:partseg}
\centering
\newcolumntype{g}{>{\columncolor{Gray}}c}
\setlength{\tabcolsep}{1.5mm}{
\begin{adjustbox}{width=\linewidth} \small
\begin{tabular}{l|c c|cccccccccccccccc}
\toprule
Methods& mIoU$_C$ & mIoU$_I$ & aero  & bag   & cap   & car   & chair & earphone & guitar & knife & lamp  & laptop & motor & mug   & pistol & rocket & skateboard  & table \\
\midrule[0.6pt]
\multicolumn{19}{c}{\textit{Supervised Representation Learning Only}}\\
\midrule[0.6pt]
PointNet~\cite{pointnet} & 80.4 & 83.7  & 83.4  & 78.7  & 82.5  & 74.9  & 89.6  & 73.0    & 91.5  & 85.9  & 80.8  & 95.3  & 65.2  & 93.0  & 81.2  & 57.9  & 72.8  & 80.6 \\
PointNet++~\cite{pointnet2} & 81.9 & 85.1  & 82.4  & 79.0  & 87.7  & 77.3  & 90.8  & 71.8  & 91.0  & 85.9  & 83.7  & 95.3  & 71.6  & 94.1  & 81.3  & 58.7  & 76.4  & 82.6 \\
DGCNN~\cite{wang2019dynamic} & 82.3 & 85.2  & 84.0    & 83.4  & 86.7  & 77.8  & 90.6  & 74.7  & 91.2  & 87.5  & 82.8  & 95.7 & 66.3  & 94.9  & 81.1  & 63.5  & 74.5  & 82.6 \\
PointMLP~\cite{pointmlp} & 84.6 & 86.1  & 83.5 & 83.4  & 87.5    & 80.5  & 90.3 & 78.2 & 92.2  & 88.1  & 82.6 & 96.2  & 77.5  & 95.8  & 85.4    & 64.6 & 83.3 & 84.3 \\
KPConv~\cite{thomas2019KPConv} & 85.1 & 86.4  & 84.6 & 86.3  & 87.2    & 81.1  & 91.1 & 77.8 & 92.6  & 88.4  & 82.7 & 96.2  & 78.1  & 95.8  & 85.4    & 69.0 & 82.0 & 83.6 \\
\midrule[0.6pt]
\multicolumn{19}{c}{\textit{Transformers with Uni-modal Generative Pre-training}}\\
\midrule[0.6pt]
Point-BERT~\cite{yu2022point} & 84.1 & 85.6  & 84.3 & 84.8  & 88.0    & 79.8  & 91.0 & 81.7 & 91.6  & 87.9  & 85.2 & 95.6  & 75.6  & 94.7  & 84.3    & 63.4 & 76.3 & 81.5 \\
Point-MAE~\cite{pang2022masked}  & 84.2  & 86.1 & 84.3 & 85.0& 88.3& 80.5& 91.3& 78.5& 92.1& 87.4& 86.1& 96.1& 75.2& 94.6& 84.7& 63.5& 77.1& 82.4\\
MaskPoint~\cite{liu2022masked} & 84.4  & 86.0 & 84.2 & 85.6& 88.1& 80.3& 91.2& 79.5& 91.9& 87.8& 86.2& 95.3& 76.9& 95.0& 85.3& 64.4& 76.9& 81.8\\
Point-M2AE~\cite{pointm2ae} & 84.9 & 86.5  & -- & --& --& --& --& --& --& --& --& --& --& --& --& --& --& -- \\
\midrule[0.6pt]
\multicolumn{19}{c}{\textit{Our Proposed 3D-to-2D Generative Pre-training}}\\
\midrule[0.6pt]
PointMLP+TAP & \textbf{85.2} & \textbf{86.9} & 84.8 & 86.1 & 89.5 & 82.5 & 92.1 & 75.9 & 92.3 & 88.7 & 85.6 & 96.5 & 79.8 & 96.0 & 85.9 & 66.2 & 78.1 & 83.2 \\
\bottomrule
\end{tabular}
\end{adjustbox}}
\end{table*}

\subsection{Downstream Tasks}
In this section, we report the experimental results of various downstream tasks. We follow the previous work to conduct experiments of object classification on real-world ScanObjectNN and synthetic ModelNet40 datasets. We also verify the effectiveness of our pre-training method on the part segmentation task with the ShapeNetPart dataset. 
\vspace{-5pt}
\subsubsection{Object Classification} 

\noindent\textbf{Main Results.}
To evaluate the effectiveness of our proposed TAP, we implement it with various point cloud architectures, including classical baselines such as PointNet++, DGCNN, and PointMLP, as well as widely used Standard Transformers backbone for existing generative pre-training methods. We follow the common practice to experiment our model on three variants of the ScanObjectNN dataset: 1) OBJ-ONLY: cropping object without any background; 2) OBJ-BG: containing the background and object; 3) PB-T50-RS: adopting various augmentations to the objects. We reported the results comparing with existing pre-training methods in Table~\ref{tab:scanobjectnn}.

\begin{figure}[t]
    \begin{center}
    \includegraphics[width=\linewidth]{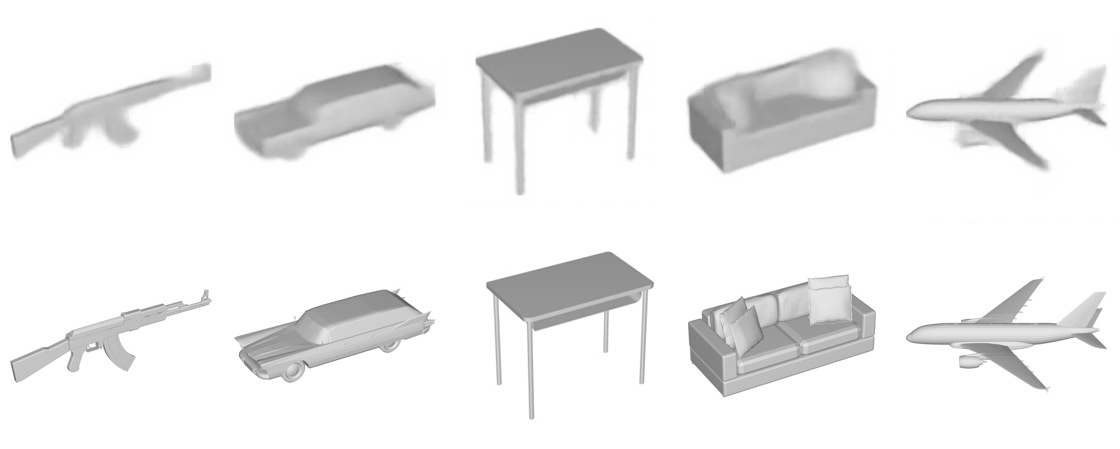}
    \caption{ \textbf{The visualization results of our proposed 3D-to-2D generative pre-training.} The first row displays view images generated by our TAP pre-training pipeline and the second row shows ground truth images. Our TAP can produce view images with appropriate shapes and reflection colors, demonstrating its ability in capturing geometric structure and stereoscopic knowledge.}
    \label{fig:examples}
    \end{center}
    \vspace{-6pt}
\end{figure}

As shown in the upper part of the table, we pre-train the graph-based architecture DGCNN, set-abstraction-based architecture PointNet++, and MLP-based architecture PointMLP with TAP, and observe consistent improvements across models. The results strongly convey that our proposed TAP can be successfully applied to various types of point cloud models and the proposed novel 3D-to-2D generative pre-training is effective regardless of the backbone architecture. Considering that nearly all existing generative pre-training methods are specially designed for Transformer-based architecture, our TAP is much superior in its wider adaptation and higher flexibility. Additionally, we also provide a detailed and fair comparison with previous work by implementing TAP with the Standard Transformers architecture in the lower part of the table, where no hierarchical designs or inductive bias is included. Our TAP outperforms previous pre-training methods in all three split settings, providing strong evidence that our 3D-to-2D generative pre-training strategy can also benefit attention-based architectures and surpass uni-modal generative pre-training competitors. It is worth noting that although TAP and many previous pre-training approaches have significantly improved the performance of Transformers in point cloud tasks, increasing the accuracy from 77.24 to 85.67, they still lag far behind advanced point cloud networks such as PointMLP. Therefore, TAP's applicability to various models is an important characteristic that would benefit future research.

\vspace{6pt}
\noindent\textbf{Comparisons with Previous Methods.}
To clearly demonstrate the high performance of our proposed TAP pre-training in object classification tasks, we compare TAP with existing methods on both synthetic ModelNet and real-world ScanObjectNN (the hardest PB-T50-RS variant) datasets in Table~\ref{tab:compare}. We categorize existing methods into two types: (1) architecture-oriented methods, which focus on developing novel model architectures for 3D point clouds and do not involve any pre-training techniques, and (2) pre-training methods, which pay more attention to the pre-training strategy and whose backbone model are mostly limited to Transformer-base architecture. It's worth noticing that methods marked with an asterisk ($*$) incorporate additional knowledge from pre-trained image models or pre-trained vision-language models. To ensure a fair and unbiased comparison, we refrain from directly comparing our method with these approaches. However, we include them in the listing for reference purposes, acknowledging their existence and potential relevance in related research.

From the experimental results, we can see that accompanied by PointMLP backbone model, our proposed TAP pre-training achieves the best classification accuracy on ScanObjectNN and ModelNet40 among existing models (with no pre-trained knowledge from image or language like P2P~\cite{wang2022p2p}), demonstrating the effectiveness of our approach and validating the superiority of our 3D-to-2D cross-modal generative pre-training method over previous generative pre-training methods. Moreover, we also note that our proposed method has brought higher performance improvements on the ScanObjectNN dataset than on ModelNet40. This may be attributed to the reason that the cross-modal generative pre-training has enhanced the network's ability to understand point clouds from different views, which is beneficial for a more robust understanding of the real-scan data with more noise and disturbance in the ScanObjectNN dataset.

\vspace{6pt}
\noindent\textbf{Visualization Results.} Figure~\ref{fig:examples} shows the visualization results of TAP. The first row shows the generated view images while the second row displays the ground truth images for reference. The TAP method can successfully predict the accurate shape of the object and the RGB colors that represent light reflections in rendered images. Therefore, TAP is capable of capturing the geometric structure of 3D objects and reasoning occlusion relations from specific camera poses.

\subsubsection{Part Segmentation} 
Performing dense prediction is always a more challenging task compared with classification. In this section, we evaluate the local distinguishability of our proposed TAP pre-training method, fine-tuning the pre-trained point cloud model on the ShapeNetPart dataset for the part segmentation task. Quantitative results are shown in Table~\ref{tab:partseg}. We implement PointMLP as the backbone model and compare our TAP results with two mainstreams of previous literature. The upper row displays classical architecture-oriented methods that focus on network design and are trained from scratch. The lower row shows members of the generative pre-training family that rely on Transformer-based architectures. 

According to results comparisons, our TAP pre-training significantly improves the part segmentation performance of the PointMLP backbone, increasing class mIoU by 0.6 and instance mIoU by 0.8. More importantly, our TAP pre-training achieves state-of-the-art performance on both class mIoU and instance mIoU, surpassing leading works in both tracks. Specifically, TAP exceeds the performance of Point-M2AE on instance mIoU by 0.4. This satisfactory performance serves as strong evidence to convey that our proposed TAP pre-training is superior to previous uni-model generative pre-training mechanisms in dense prediction tasks. This may be attributed to the factor that our supervision in 2D with MSE loss is more precise than the ambiguous Chamfer Distance in 3D reconstruction. Therefore, models with TAP pre-training obtain more accurate comprehension of local geometry and detail awareness, which contributes to mIoU gain in dense prediction tasks. What's more, TAP outperforms KPConv on instance mIoU by 0.5, demonstrating that the proposed 3D-to-2D generative pre-training method can fully exploit the potential of the point cloud model and help it have a better perception of objects' geometric structure. As TAP is adaptable to any architecture, future improvements in architectural design can also benefit from TAP pre-training.

\subsubsection{Few-shot Classification}
Following Point-BERT~\cite{yu2022point}, we conduct few-shot classification with Standard Transformers on ModelNet40~\cite{modelnet} dataset. As shown in Table~\ref{tab:fewshot}, we report mean overall accuracy and standard deviation (mOA$\pm$std) on 10 pre-defined data folders for each few-shot setting. The \textit{way} and \textit{shot} in Table~\ref{tab:fewshot} specify the number of categories and the number of training examples per category, respectively. 

From the results, TAP achieves the highest mean overall accuracy across all few-shot settings when compared to previous generative pre-training approaches. Furthermore, TAP exhibits significantly lower standard deviations than those reported in the existing literature for the majority of few-shot settings, which signifies its robust performance and consistent superiority. This indicates that TAP is not only capable of achieving high mean overall accuracy but also exhibits reliability and robustness across various few-shot settings. Such stability is crucial in real-world applications, where consistency and predictability are vital for practical deployment.

\subsection{Scene-level Dense Predictions}
To assess the effectiveness of TAP in handling scene-level dense prediction tasks, we carry out experiments on more complicated scene-level object detection and semantic segmentation on the ScanNetV2~\cite{dai2017scannet} dataset. For the object detection task, we adopt 3DETR~\cite{misra20213detr} and pre-train its encoder on the object-level dataset ShapeNet~\cite{shapenet} with TAP. Average precision at 0.25 and 0.5 IoU thresholds are reported. Regarding semantic segmentation, we employ the PointTransformerV2~\cite{wu2022ptv2}(PTv2) model and pre-train it on the ScanNetV2 dataset with TAP. We report mean IoU for evaluation metric. It is worth mentioning that PTv2 represents the current state-of-the-art approach with open-source code availability.

\begin{table}[t]
    \centering
    \caption{\textbf{Few-shot Classification with Standard Transformers on ModelNet40 dataset.} We report mean overall accuracy and standard deviation on 10 pre-defined data folders for each setting. Best results are marked \textbf{bold}.}
    \label{tab:fewshot}
    \setlength\tabcolsep{2pt}
    \resizebox{\linewidth}{!}{
    \begin{tabular}{lcccc}
    \toprule[0.95pt]
    \multirow{2}{*}[-0.5ex]{Method}& \multicolumn{2}{c}{5-way} & \multicolumn{2}{c}{10-way}\\
    \cmidrule(lr){2-3}\cmidrule(lr){4-5} & 10-shot & 20-shot & 10-shot & 20-shot\\
    \midrule[0.6pt]
     w/o pre-training & 87.8 $\pm$ 5.2& 93.3 $\pm$ 4.3 & 84.6 $\pm$ 5.5 & 89.4 $\pm$ 6.3\\
     \midrule
     Point-BERT~\cite{yu2022point} & 94.6 $\pm$ 3.1 & 96.3 $\pm$ 2.7 &  91.0 $\pm$ 5.4 & 92.7 $\pm$ 5.1\\
     MaskPoint~\cite{liu2022masked} & 95.0 $\pm$ 3.7 & 97.2 $\pm$ \textbf{1.7} & 91.4 $\pm$ 4.0 & 93.4 $\pm$ 3.5\\
     Point-MAE~\cite{pang2022masked} & 96.3 $\pm$ 2.5 & \textbf{97.8} $\pm$ 1.8 & 92.6 $\pm$ 4.1 & 95.0 $\pm$ 3.0\\
     \midrule
     TAP   & \textbf{97.3} $\pm$ \textbf{1.8} & \textbf{97.8} $\pm$ \textbf{1.7} &  \textbf{93.1} $\pm$ \textbf{2.6} & \textbf{95.8} $\pm$ \textbf{1.0} \\
     \bottomrule[0.95pt]
     \end{tabular}
    }
\end{table}

\begin{table}[!t]
    \centering
    \caption{\textbf{Scene-level object detection and semantic segmentation on ScanNetV2~\cite{dai2017scannet}}. Average precision at 0.25 IoU thresholds (AP$_\textrm{0.25}$) and 0.5 IoU thresholds (AP$_\textrm{0.5}$) of detection and mean Intersection-over-Union (mIoU) of semantic segmentation are reported.}
    \label{tab:scene}
    \resizebox{\linewidth}{!}{
    \begin{tabular}{lccc}
    \toprule[0.95pt]
    \multirow{2}{*}[-0.5ex]{Method}& \multicolumn{2}{c}{Det (3DETR~\cite{misra20213detr})} & Seg (PTv2~\cite{wu2022ptv2}) \\
    \cmidrule(l){2-3}\cmidrule(l){4-4} 
    & AP$_\textrm{0.25}$ & AP$_\textrm{0.5}$ & mIoU \\
    \midrule[0.6pt]
     Baseline & 62.1 & 37.9 & 72.4 \\
     +TAP & 63.0~\cred{(+0.9)} & 41.4~\cred{(+3.5)} & 72.6~\cred{(+0.2)} \\
     \bottomrule[0.95pt]
     \end{tabular}}
\end{table}

Based on the results presented in Table~\ref{tab:scene}, TAP consistently enhances the performance of all baselines, thereby showcasing its efficacy in tackling more intricate scene-level dense prediction tasks. Remarkably, even with the encoder solely pre-trained on an object-level dataset for scene-level detection task, significant improvements are observed in both AP$_\textrm{0.25}$ and AP$_\textrm{0.5}$ metrics. This suggests that the learned representations from TAP effectively capture relevant information and generalize well to complex scenes, even when the pre-training data is limited to object-level collections. Such generalization capabilities are valuable in scenarios where obtaining large-scale fully annotated scene-level datasets may be challenging or expensive.

\begin{table*}[!t]
\label{tab:ablation}
\caption{\small \textbf{Ablation studies on Photograph module in TAP pre-training pipeline.} We choose PointMLP as the backbone model and conduct ablation studies on the ScanObjectNN dataset from two aspects: overall architectural designs and query designs. In Table(a), we first investigate the effectiveness of cross-attention design compared with direct projection (Model A$_1$) and direct project with self-attention (Model A$_2$). Then we analyze the influence of the different number of attention layers and feature channels in Model B and Model C. Finally, we discuss whether pad token in memory builder is beneficial in Model D. In Table(b), we conduct further experiments to compare different approaches for query designing: (1) Using learnable query based on the given viewpoints (Model E) or use mathematical formulations derived in Eq.~\ref{eq:line}. (2) The information we need when we mathematically encode pose information into init query status: (i) \textbf{Origin}: the coordinate of origin point $O$ that the optical line passes through. (ii) \textbf{Direction}: The normalized direction of the optical line. (iii) \textbf{PE}: The position embedding for each grid.}
\vspace{-6pt}
\centering
\begin{minipage}{.54\textwidth}
\begin{subtable}{\textwidth}
    \setlength\tabcolsep{3pt}
    \caption{Overall Architectural Designs.}
    \vspace{-5pt}
    \centering
    \label{tab:abl_arch}
    \newcolumntype{g}{>{\columncolor{Gray}}c}
    \adjustbox{width=\textwidth}{
    \begin{tabular}{@{\hskip 3pt}>{\columncolor{white}[3pt][\tabcolsep]}c|c|ccc| >{\columncolor{Gray}[\tabcolsep][3pt]}g@{\hskip 3pt}}
    \toprule
        Model     & Attention Type  & LayerNum. &  Channels & Mem.Pad  &   Acc.(\%) \\
    \midrule
        A$_1$      & None  &  -- &   256   & \xmark    & 87.6~\cb{(-0.9)} \\
        A$_2$        & SelfAttn  &  6 layers &  256     & \xmark      & 87.8~\cb{(-0.7)} \\
    \midrule
        B       & CrossAttn  & 2 layers &    256  & \cmark      & 87.9~\cb{(-0.6)} \\
        C       & CrossAttn  & 6 layers&   512   & \cmark      & 87.8~\cb{(-0.7)} \\
        D        & CrossAttn &  6 layers&   256      & \xmark     & 88.3~\cb{(-0.2)} \\
    \midrule
        TAP      & CrossAttn & 6 layers &  256    & \cmark       & 88.5 \\ 
    \bottomrule
    \end{tabular}}
\end{subtable}

\end{minipage} \hfill
\begin{minipage}{.44\textwidth}
\begin{subtable}{\textwidth}
    \setlength\tabcolsep{3pt}
    \caption{Query Designs.}
    \vspace{-5pt}
    \label{tab:abl_query}
    \newcolumntype{g}{>{\columncolor{Gray}}c}
    \adjustbox{width=\textwidth}{
    \begin{tabular}{@{\hskip 3pt}>{\columncolor{white}[3pt][\tabcolsep]}c|c|ccc |>{\columncolor{Gray}[\tabcolsep][3pt]}g@{\hskip 3pt}}
    \toprule
        Model   & Query Type  & Origin & Direction & PE   & Acc.(\%) \\
    \midrule
        E       & Learnable    & \xmark      & \xmark      &  \xmark & 87.5~\cb{(-1.0)}        \\
    \midrule
        F$_1$   & Formula   & \xmark      & \cmark     & \cmark  &  86.5~\cb{(-2.0)} \\
        F$_2$    & Formula    & \cmark     & \xmark      &  \xmark & 87.8~\cb{(-0.7)}       \\
        F$_3$   & Formula   & \cmark     & \cmark     & \xmark  & 88.0~\cb{(-0.5)}        \\
        F$_4$   & Formula   & \cmark      & \xmark     & \cmark  & 88.1~\cb{(-0.4)}    \\
    \midrule
        TAP     & Formula   & \cmark    & \cmark      & \cmark  & 88.5        \\
    \bottomrule
    \end{tabular}}
\end{subtable}
\end{minipage}
\end{table*}

\subsection{Ablation Studies}

To investigate the architectural design of our proposed \textit{Photograph Module} in TAP pre-training pipeline, we conduct extensive ablation studies on the ScanObjectNN dataset with PointMLP as the backbone model.

\vspace{6pt}
\noindent\textbf{Photograph Module Architectural Designs.}
In Photograph Module, we implement cross-attention layers to generate view image feature maps conditioned on pose instruction. We believe that letting the module learn by itself how to rearrange 3D point features in 2D grids will enhance the representation ability of the 3D backbone. Therefore, we conduct ablation studies to verify this hypothesis. As shown in Table~\ref{tab:abl_arch}, we implement Model A$_1$ with no attention layers, directly projecting 3D feature points to 2D grids based on Eq.~\ref{eq:rotate} and Eq.~\ref{eq:proj}. This results in a much simpler pre-training task, as the projection relation has been directly told. Additionally, in Model A$_2$, we add self-attention layers after explicit projection to help the model capture longer-range correlations. Pose knowledge is encoded as a pose token that is concatenated to projected grids, similar to the CLS token in classification Transformers. According to quantitative results comparison with TAP that implements cross-attention layers, fine-tuning results of pre-training methods in A$_1$ and A$_2$ version show inferiority. Therefore, the cross-attention architecture we designed to entirely LEARN the projection relation is the most suitable choice for the proposed 3D-to-2D generative pre-training.

What's more, we discuss the number of cross-attention layers, the dimension of feature channels and whether to concatenate pad token in memory builder in Model $B,C,D$. According to the results, more cross-attention layers show stronger representation ability, while too large channel number will lead to performance decrease caused by over-fitting. The performance gain from Model $D$ to TAP also verifies that the pad token design in the memory builder is essential.

\vspace{6pt}
\noindent\textbf{Query Generator Designs.} In the query generator, we derive the mathematical formulation of the optical lines passing through 2D grids. We propose to concatenate the coordinate of origin point $O$, normalized direction $\mathbf{d}^\dagger$ and position embedding $(u/h, v/w)$ as the initial state of queries. In Table~\ref{tab:abl_query}, we first compare this mathematical design with totally learnable queries that takes pose matrix $R$ as input and implements MLP layers to predict query for each grid. As shown in Model $E$, learnable queries cannot satisfactorily encode pose information, while our derived formulation for query construction is both clearer in physical meaning and more competitive in fine-tuning accuracy.

In ablation F$_1$ to F$_4$, we progressively discuss the three components of query generation. Quantitative comparison with TAP verifies that every component is indispensable for query generation, where coordinates of origin points are of the most importance. 

\vspace{6pt}
\section{Conclusions}

In this paper, we have proposed a novel 3D-to-2D generative pre-training method TAP that is adaptable to any point cloud model. We implemented the cross-attention mechanism to generate view images of point clouds from instructed camera poses. To better encode pose conditions and generate physically meaningful queries, we derived mathematical formulations of optical lines. The proposed TAP pre-training had higher preciseness in supervision and broader adaptation to different backbones, compared with directly reconstructing point clouds in previous methods. Experimental results conveyed that the TAP pre-training can help the backbone models better capture the structural knowledge and stereoscopic relations. Fine-tuning results of TAP pre-training achieve state-of-the-art performance on ScanObjectNN classification and ShapeNetPart segmentation, among methods that do not include any pre-trained image or text models. We believe the cross-modal generative pre-training paradigm will be a promising direction for future research.

\section*{Acknowledgement}
This work was supported in part by the National Key Research and Development Program of China under Grant 2022ZD0114903 and in part by the National Natural Science Foundation of China under Grant 62125603.

\newpage
\clearpage
{\small
\bibliographystyle{ieee_fullname}
\bibliography{ref}
}

\newpage
\clearpage
\appendix

\section{Ablation Studies}
In this section, we conduct more ablation studies on hyperparameter choices of the proposed 3D-to-2D generative pre-training, discussing more thoroughly the insights into architectural design and objective function design. We implement PointMLP~\cite{pointmlp} as the 3D backbone model and conduct these ablation experiments on the hardest PB-T50-RS variant of the ScanObjectNN~\cite{uy2019revisiting} dataset. We report the classification accuracy of the fine-tuning results.

\subsection{Cross-Attention Hyperparameters}

In Table~\ref{tab:abl_attn}, we display the results of ablation studies on the number of layers and feature channels of the cross-attention layers in our proposed Photograph module. From the quantitative results, we can conclude that 2 layers with 128 channels is the best hyperparameter group for cross-attention layers. When we implement a shallow layer setting (2 layers in line 1 and 4 layers in line 2), lower feature channels (128 dims) achieves better performance. On the contrary, when we implement a deeper layer setting (6 layers in line 3 and 8 layers in line 4), relatively higher feature channels (256 dims) is the best choice. Additionally, if we use 1024 dims as the feature channels in cross-attention layers, which is the same as the channels of output features from the 3D backbone model, the pre-training stage totally collapses and the fine-tuning results are much lower than models of 128 dims and 256 dims, no matter how much layers are implemented. This result indicates that a bottleneck design in our proposed photograph module is essential for the successful pre-training of the proposed 3D-to-2D generation.

The overall trend is that a lightweight architectural design of the cross-attention layers is better than a heavy module design. This may be because we completely drop the photograph module and only keep the 3D backbone in the fine-tuning stage. Therefore, a lightweight photograph module in the pre-training stage will encourage the 3D backbone to exploit more representation ability and avoid information loss in the fine-tuning stage to the best extent. On the contrary, if we implement a heavy photograph module with deep cross-attention layers and high feature dimensions, the photograph module will dominate the generation process and the importance of the 3D backbone will be neglected. What's worse, in the fine-tuning stage, the rich geometry information in the heavy photograph module is totally dropped out and no longer helpful for downstream tasks.

\begin{table}[!t]
\label{tab:ablation_supp}
\caption{\textbf{Ablation Studies on Hyperparameters.} We implement PoinMLP~\cite{pointmlp} as the 3D backbone model and conduct experiments on the hardest PB-T50-RS variant of ScanObjectNN~\cite{uy2019revisiting} dataset.}
\vspace{-6pt}
\centering
\begin{subtable}{0.49\textwidth}
    \setlength\tabcolsep{3pt}
    \centering
    \caption{Cross-Attention Hyperparameters.}
    \vspace{-4pt}
    \label{tab:abl_attn}
    \adjustbox{width=0.9\textwidth}{
    \begin{tabular}{c|ccc}
    \toprule
    LayerNum $\backslash$ Channels & 128 Dims  & 256 Dims &  1024 Dims \\
    \midrule
    2 Layers    & \textbf{89.1} & 87.9 & 86.3 \\
    4 Layers    & 88.7 & 88.0 & 85.9 \\
    6 Layers    & 88.3 & 88.5 & 85.3 \\
    8 Layers    & 87.7 & 88.1 & 85.8 \\
    \bottomrule
    \end{tabular}}
\end{subtable}
\hfill
\vspace{5pt}
\begin{subtable}{0.49\textwidth}
    \centering
    \setlength\tabcolsep{3pt}
    \caption{Loss Weight Hyperparameters.}
    \vspace{-4pt}
    \label{tab:abl_lossw}
    \newcolumntype{g}{>{\columncolor{Gray}}c}
    \adjustbox{width=0.9\textwidth}{
    \begin{tabular}{@{\hskip 3pt}>{\columncolor{white}[3pt][\tabcolsep]}c|cccccc|cc}
    \toprule
        Model   & G$_1$ & G$_2$ & G$_3$ & G$_4$ & G$_5$ & G$_6$ & H$_1$ & H$_2$ \\
    \midrule
        $w^\textrm{fg}$ & 2 & 5 & 10 & 20 & 30 & 50 & 0 & 20\\
        $w^\textrm{bg}$ & 1 & 1 & 1 & 1 & 1 & 1 & 0 & 1\\
        $w^\textrm{feat}$ & 0 & 0 & 0 & 0 & 0 & 0 & 2 & 2\\
    \midrule
        Acc. (\%) & 87.1 & 87.2 & 88.0 & \textbf{88.5} & 88.0 & 86.8 & 86.3 & 87.8 \\
    \bottomrule
    \end{tabular}}
\end{subtable}
\end{table}

\begin{figure*}[!t]
     \centering    
     \includegraphics[width=0.95\textwidth]{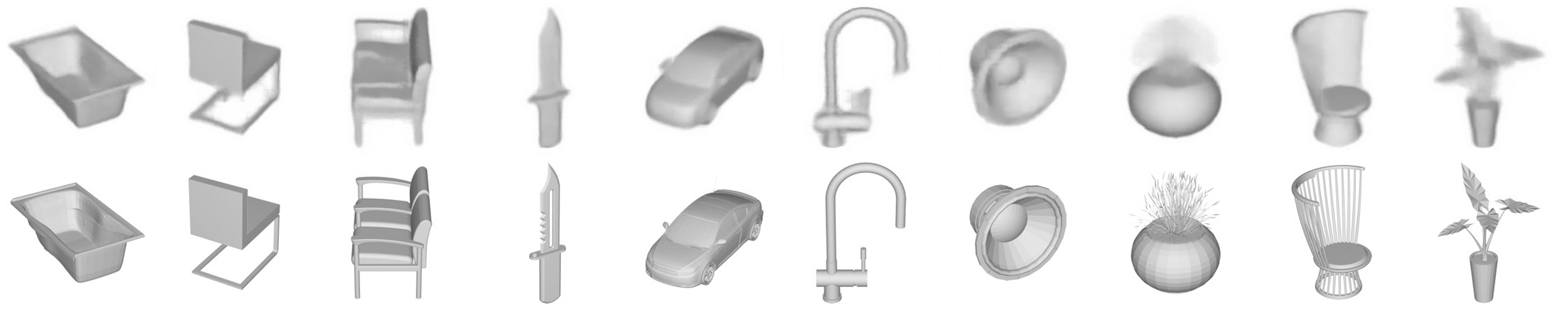}
     \caption{Visualization of the outputs from the 3D-to-2D generative pre-training. The first line shows the generated view images from the model. The second line shows the ground truth images for reference.}
     \label{fig:examples_supp}
\end{figure*}

\subsection{Objective Function}
In this subsection, we discuss the objective function design of our proposed 3D-to-2D generative pre-training. In our main paper, we implement pixel-level supervision with MSE loss between generative view images $I_\textrm{gen}$ and ground truth images $I_\textrm{gt}$:
\begin{equation}
    \mathcal{L}_\textrm{pix}(I_\textrm{gen}, I_\textrm{gt}) = w^\textrm{fg} \mathcal{D}(I_\textrm{gen}^\textrm{fg}, I_\textrm{gt}^\textrm{fg}) + w^\textrm{bg} \mathcal{D}(I_\textrm{gen}^\textrm{bg}, I_\textrm{gt}^\textrm{bg})
\end{equation}
where fg denotes foreground region, bg denotes background region and $\mathcal{D}$ is the MSE distance. However, in 2D generation, perceptual loss~\cite{johnson2016perceptual} is of equal importance with pixel-wise loss. While pixel-wise MSE loss focuses on low-level similarities, perceptual loss measures high-level semantic differences between feature representations of the images computed by the pre-trained loss network. Technically, perceptual loss makes use of a loss network $\phi$ pre-trained for image classification, which is typically a 16-layer VGG~\cite{simonyan2014vgg} network pre-trained on the ImageNet~\cite{russakovsky2015imagenet} dataset. If we denote $\phi_j(x)$ as the feature map with size $c_j\times h_j\times w_j$ of the $j$th layer of the network $\phi$, then the perceptual loss is defined as the Euclidean distance:
\begin{equation}
\small
    \mathcal{L}_\textrm{feat}(I_\textrm{gen}, I_\textrm{gt}) = \frac{1}{N}\sum_j\frac{1}{c_j h_j w_j}\lVert\phi_j(I_\textrm{gen})- \phi_j(I_\textrm{gt})\rVert_2^2
\end{equation}
where $N$ is the number of total layers of the VGG network and $1\leq j\leq N$. If we combine the pixel-wise loss $\mathcal{L}_\textrm{pix}$ with the perceptual loss $\mathcal{L}_\textrm{feat}$, then the final objective function of the proposed 3D-to-2D generation is:
\begin{equation}
\small
    \mathcal{L} = \mathcal{L}_\textrm{pix} + w^\textrm{feat} \mathcal{L}_\textrm{feat}
\end{equation}

In Table~\ref{tab:abl_lossw}, we conduct ablations on loss weight of foreground pixel-wise loss $w^\textrm{fg}$, background pixel-wise loss $w^\textrm{bg}$ and perceptual loss $w^\textrm{feat}$. In Model $G_1$ to $G_6$, we only implement pixel-wise loss. In Model $H_1$, we only implement perceptual loss. In Model $H_2$, we combine pixel-wise loss with perceptual loss. From the ablation results, we can conclude that $w^\textrm{fg}:w^\textrm{bg}=20:1$ is the best hyperparameter choice for pixel-wise loss. However, the perceptual loss is not effective when we compare Model $G_4$, Model $H_1$ and Model $H_2$. This is mainly due to the reason that the rendered view image of synthetic ShapeNet~\cite{shapenet} dataset is out of the distribution of the realistic ImageNet~\cite{russakovsky2015imagenet} that the loss model $\phi$ is pre-trained on. Therefore, the high-level semantic representation ability of $\phi$ on view images is relatively poor and cannot guide the optimization of the 3D-to-2D generation process. If the rendered images are more realistic with colors and background, then the perceptual loss is expected to help 3D-to-2D generative pre-training.

\begin{figure}[t]
     \centering
     \begin{subfigure}{0.23\textwidth}
         \centering
         \includegraphics[width=0.95\textwidth]{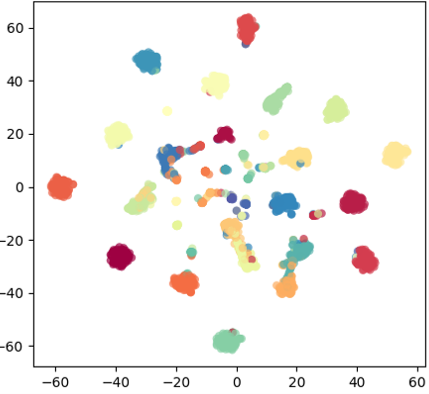}
         \caption{ModelNet40.}
         \label{fig:tsne_modelnet}
     \end{subfigure}
     \begin{subfigure}{0.23\textwidth}
         \centering
         \includegraphics[width=0.95\textwidth]{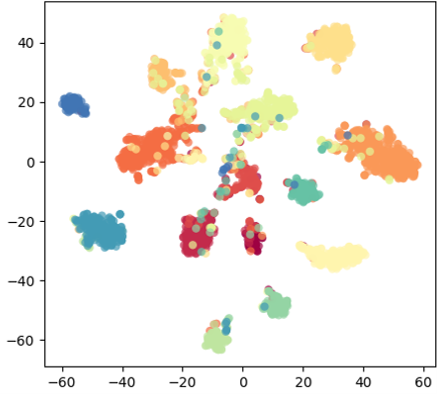}
         \caption{ScanObjectNN.}
         \label{fig:tsne_scanobjectnn}
     \end{subfigure}
     \vspace{-5pt}
     \caption{Visualization of feature distributions in t-SNE representations. Best view in colors.}
     \label{fig:tsne}
\end{figure}

\section{Visualization Results}

\subsection{Generated View Images}

Figure~\ref{fig:examples_supp} displays more visualization results of our generated view images from the 3D-to-2D generative pre-training process. We take ShapeNet~\cite{shapenet} as the pre-training dataset and implement PointMLP~\cite{pointmlp} as the 3D backbone model. The first line shows the generated results from our model while the second line shows ground truth images for reference. The visualization results convey that our 3D-to-2D generative pre-training can successfully predict the shape and colors of the objects from specific projection views. There are also some unsatisfactory cases in the last three columns, where there are some vague details in our generated images. This is mainly due to the large downsample ratio ($\times32$) in our model design.

\subsection{Feature Distributions}

Figure~\ref{fig:tsne} shows feature distributions of ModelNet40~\cite{modelnet} and ScanObjectNN~\cite{uy2019revisiting} datasets in t-SNE visualization. We choose PointMLP~\cite{pointmlp} as the 3D backbone and pre-train on ShapeNet~\cite{shapenet} dataset. We can conclude that with our proposed 3D-to-2D pre-training, the 3D backbone model can extract discriminative features after fine-tuning on downstream classification datasets.

\begin{figure}[t]
     \centering
     \includegraphics[width=0.48\textwidth]{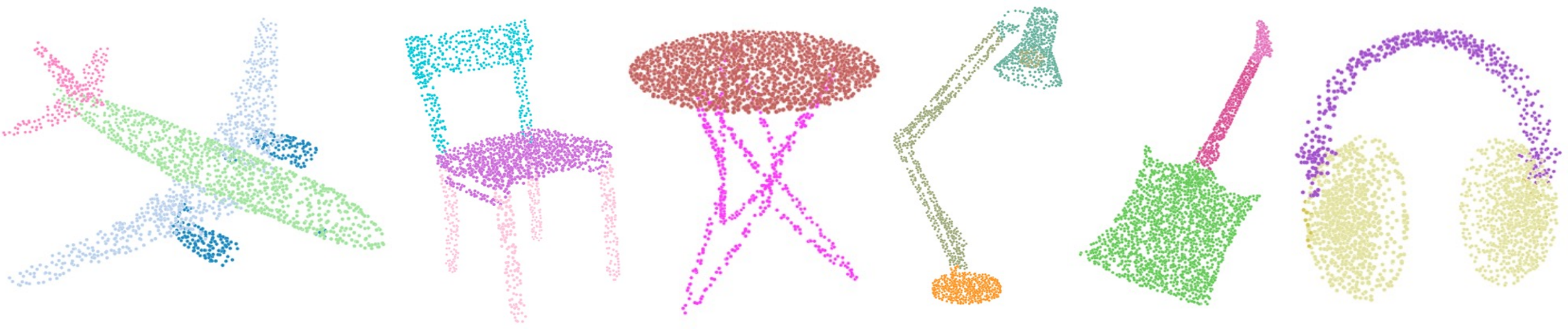}
     \caption{Illustration of part segmentation results.}
     \label{fig:partseg}
\end{figure}

\subsection{Part Segmentation Visualizations}

Figure~\ref{fig:partseg} presents visualizations of part segmentation results on samples from the ShapeNetPart dataset. Each part is represented by a distinct color for clarity. These qualitative results serve as compelling visual evidence and provide a vivid illustration of the efficacy of our fine-tune model in achieving accurate part segmentation.

\end{document}